%% file: diagnosis_and_prognosis.tex
\documentclass[a4paper, twoside, 12pt]{article}
\input{template}
\usepackage{subcaption}

\begin{document}

\newcommand{\runningauthor}{Venkatraghavan \textit{et al.}} 

\newcommand{\runningheadtitle}{Computer-aided diagnosis and prediction in brain disorders}

\newcommand{\chapternumber}{15}

\newcommand{\emailaddress}{e.bron@erasmusmc.nl}

\title{Computer-aided diagnosis and prediction in brain disorders} 

\author[1,2]{Vikram Venkatraghavan}
\author[3]{Sebastian R. van der Voort}
\author[4,5]{Daniel Bos}
\author[4]{Marion Smits}
\author[6,7]{Frederik Barkhof}
\author[3,8]{Wiro J. Niessen}
\author[3]{Stefan Klein $^\dagger$}
\author[*,3]{Esther E. Bron $^\dagger$}

\affil[1]{Alzheimer Center Amsterdam, Neurology, Vrije Universiteit, Amsterdam UMC location VUmc, Amsterdam, the Netherlands}
\affil[2]{Amsterdam Neuroscience, Neurodegeneration, Amsterdam, the Netherlands}
\affil[3]{Biomedical Imaging Group Rotterdam, Department of Radiology and Nuclear Medicine, Erasmus MC, the Netherlands}
\affil[4]{Department of Radiology and Nuclear Medicine, Erasmus MC, the Netherlands}
\affil[5]{Department of Epidemiology, Erasmus MC, the Netherlands}
\affil[6]{Department of Radiology and Nuclear Medicine, Amsterdam UMC, the Netherlands}
\affil[7]{Insititutes of Neurology and Healthcare Engineering, University College London, UK}
\affil[8]{Quantitative Imaging Group, Dept. of Imaging Physics, Faculty of Applied Sciences, TU Delft, the Netherlands}

\affil[*]{Corresponding author: e-mail address: \href{mailto:e.bron@erasmusmc.nl}{\emailaddress}}
\affil[$^\dagger$]{Authors contributed equally to this chapter.}

\maketitle

\afterpage{\aftergroup\restoregeometry}
\pagestyle{otherpages}

\begin{abstract}

Computer-aided methods have shown added value for diagnosing and predicting brain disorders and can thus support decision making in clinical care and treatment planning. This chapter will provide insight into the type of methods, their working, their input data - such as cognitive tests, imaging and genetic data - and the types of output they provide. We will focus on specific use cases for diagnosis, i.e. estimating the current ‘condition’ of the patient, such as early detection and diagnosis of dementia, differential diagnosis of brain tumours, and decision making in stroke. Regarding prediction, i.e. estimation of the future ‘condition’ of the patient, we will zoom in on use cases such as predicting the disease course in multiple sclerosis and predicting patient outcomes after treatment in brain cancer. Furthermore, based on these use cases, we will assess the current state-of-the-art methodology and highlight current efforts on benchmarking of these methods and the importance of open science therein. Finally, we assess the current clinical impact of computer-aided methods  and discuss the required next steps to increase clinical impact.

\end{abstract}


\section{Introduction} \label{sec:intro}

Computer-aided methods have major potential value for diagnosing and predicting outcomes in brain disorders such as dementia, brain cancer, and stroke. Diagnosis aims to determine the current ‘condition’ of the patient. Prediction, or prognosis, on the other hand, aims to forecast the future `condition' of the patient. In this way, the patient’s current and future condition can be estimated in a more detailed and accurate way, which opens up possibilities for better patient care and personalised medicine, with interventions tailored to the individual patient. Moreover, diagnosis and prediction are not only crucial for decision making in clinical care and treatment planning, but also for managing the expectations of patients and their caregivers. This is particularly important in brain disorders as they may strongly affect life expectancy and quality of life, as symptoms of the disorder and side effects of the treatment can have a major impact on the patient’s cognitive skills, daily functioning, social interaction, and general well-being. In clinical practice, diagnosis and prediction are typically performed using multiple sources of information, such as symptomatology, medical history, cognitive tests, brain imaging, electroencephalography (EEG), magnetoencephalography (MEG), blood tests, cerebrospinal fluid (CSF) biomarkers, histopathological or molecular findings, and lifestyle and genetic risk factors. These various pieces of information are integrated by the treating clinician, often in consensus with other experts at a multidisciplinary team meeting, in order to reach a final diagnosis and/or treatment plan. The aim of computer-aided methods for diagnosis and prediction is to support this process, in order to achieve more accurate, objective, and efficient decision making.

In the literature, numerous examples of computer-aided methods for diagnosis and prediction in brain disorders can be found. Most of the state-of-the-art methods use some form of machine learning to construct a model that maps (often high-dimensional) input data to the output variable of interest. There exists a large variation in machine learning technology, types of input data, and output variables. Chapters 1-6 introduced the main machine learning technologies used for computer-aided diagnosis and prediction. These include, on the one hand, classical methods such as linear models, support vector machines, and random forests, and on the other hand deep learning methods such as convolutional neural networks and recurrent neural networks. {These methods can either be implemented as classification models (estimating discrete labels) or as regression models (estimating continuous quantities), possibly specialized for survival (or `time-to-event’) analysis. In addition, Chapter 17 highlights the category of disease progression modelling techniques, which could be considered as a specialized form of machine learning incorporating models of the disease evolution over time.} Chapters 7-12 described the main types of input data used in machine learning for brain disorders: clinical evaluations, neuroimaging, EEG/MEG, genetics and omics data, electronic health records, smartphone and sensor data. The current chapter focuses on the choice of the output variable, i.e., the diagnosis or prediction of interest.

\begin{figure}[t]
\centering
\includegraphics[width=\textwidth]{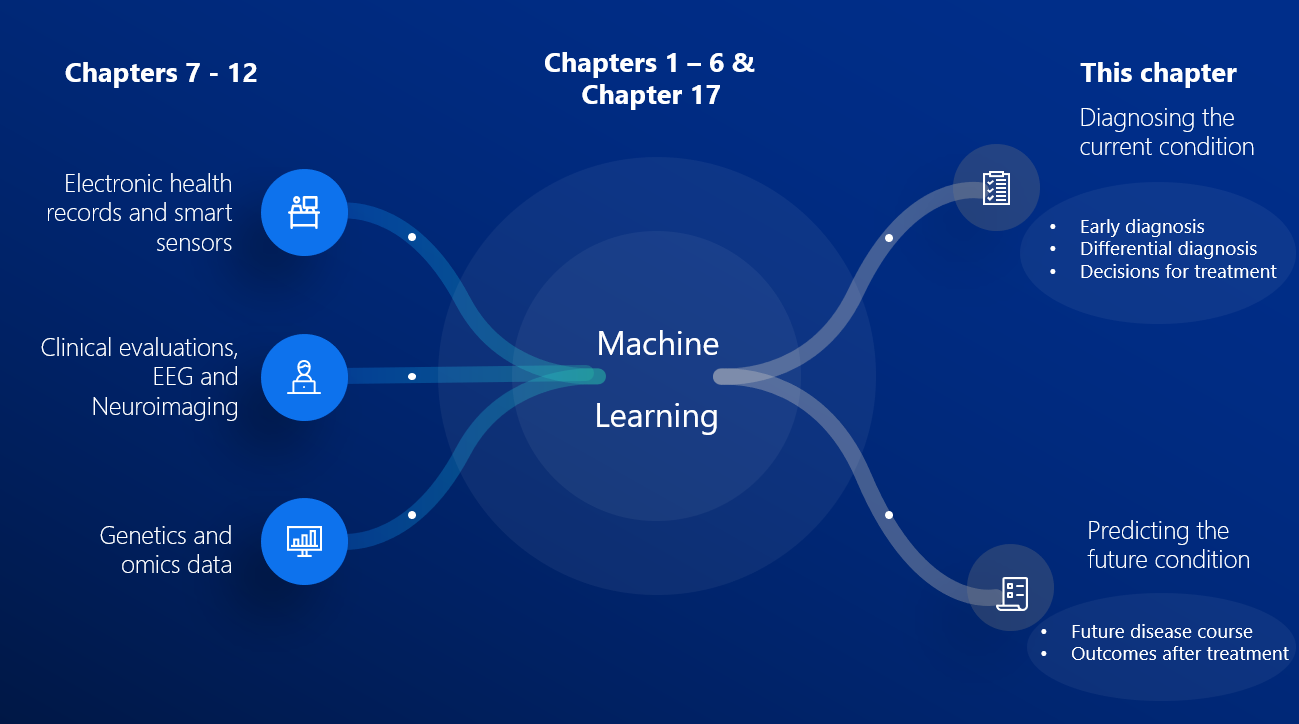}
\caption{Overview of the topics covered in this chapter, in the context of the other chapters in this book.}
\label{fig:illustration}
\end{figure}

To illustrate the various ways in which machine learning could aid diagnosis and prediction, we focus on representative use cases organised according to the type of output. Section \ref{sec:introDiagnosis} presents diagnostic use cases, including early diagnosis, differential diagnosis, and decision making for treatment. Section \ref{sec:introPrediction} presents prediction use cases, including estimation of the natural disease course and prediction of patient outcomes after treatment. While the diagnostic use cases are the core of current clinical practice which could be aided by machine learning, the prediction use cases represent a potential future application. Currently, prediction is not so often made as clinicians are not yet able to make a reliable prediction in most cases. After these introductory sections, Section \ref{sec:method} provides a more comprehensive survey of the state-of-the-art methodology, and Section \ref{sec:impact} analyses the clinical impact of such methodology and suggests a roadmap for further clinical translation. Finally, Section \ref{sec:conclusion} concludes this chapter.

\subsection{Diagnosis} \label{sec:introDiagnosis}

Diagnosis aims to determine the current ‘condition’ of the patient to inform patient care and treatment decisions. Here, we introduce three categories of diagnostic tasks that occur in clinical practice, and describe why and how computer-aided models have or could have added value. 

\begin{floatbox}[H]
    \begin{nicebox}[Diagnosis]
        \label{box:diagnosis}
         Categories of diagnostic tasks that occur in clinical practice in which computer-aided models have or could have added value, with brain disorders for which this is relevant as examples:
        \begin{itemize}
            \item \textbf{Early diagnosis} Dementia, MS
            \item \textbf{Differential diagnosis} Dementia, brain cancer
	\item \textbf{Decision making for treatment} Stroke
        \end{itemize}
    \end{nicebox}
\end{floatbox}

\textbf{Early diagnosis} is highly challenging in neurodegenerative diseases such as dementia and multiple sclerosis (MS). Dementia is a clinical syndrome which can be caused by several underlying diseases, Alzheimer’s disease (AD) being the most prevalent, and is estimated to affect 50 million people worldwide \citep{AlzheimersAssociation2020}. The mean age at dementia diagnosis is approximately 83 years~\citep{Plassman2011}. MS is estimated to affect about 2 million people worldwide, and it primarily affects younger adults with the mean age of onset for incident MS being approximately 30 years~\citep{Liguori2000}. Both for dementia and MS, establishing the diagnosis usually takes a substantial period of time after the first clinical symptoms arise \citep{VanVliet2013,Kaufmann2018}.  Early detection and accurate diagnosis is crucial for timely decision making regarding care and management of dementia symptoms, and as such can reduce healthcare costs and improve quality of life as it gives patients access to supportive therapies that help to delay institutionalisation \citep{Prince2011}. Early diagnosis of MS is important, because patients who begin treatment earlier do reap more benefit than those who start late \citep{Miller2004}. In addition, advancing the diagnosis in time is essential to support the development of new disease modifying treatments, since late treatment is expected to be a major factor in failure of clinical trials \citep{Mehta2017}. The clinical diagnosis of dementia is currently based on objective assessment of cognitive impairment, {assessment of biomarkers} \citep{Dubois2014} and evaluation of its interference with daily living \citep{McKhann2011, Albert2011, Gorelick2011, Rascovsky2011}. The clinical diagnosis of MS is based on frequency of relapsing inflammatory attacks, associated symptoms and distribution of lesions on MRI \citep{Thompson2018}. {For a subset of MS patients with demyelinating lesions highly suggestive of MS, termed as radiologically isolated syndrome (RIS), a separate diagnostic criteria was formed by Okuda \textit{et al.}}\citep{Okuda2014} {to improve the diagnostic accuracy.} However, objective assessment of biomarkers of the underlying processes can advance diagnosis, since symptoms are known to arise relatively late in the disease process. This holds for example for cognitive impairment due to dementia and physical disability or cognitive impairment due to MS \citep{Jack2013, Gordon2018, Dekker2021}. By combining neuroimaging and other biomarkers with machine learning based on large-datasets, computer-aided diagnosis algorithms aim to facilitate medical decision support by providing a potentially more objective diagnosis than that obtained by conventional clinical criteria \citep{Kloppel2012, Rathore2017}. In addition to biomarkers, machine learning based on data from remote monitoring technology, such as wearables and smart watches, is an emerging field of research aimed at detecting cognitive, behavioural and physical symptoms in an objective way at the earliest stage possible \citep{Muurling2021, Simblett2020}. 

Beyond an early diagnosis, accurate identification of the underlying disease, i.e. \textbf{differential diagnosis}, is crucial for planning care and treatment decisions. For example, in dementia, the most common underlying diseases are AD, vascular cognitive impairment (VCI), {dementia with Lewy bodies (DLB)}, and fronto-temporal lobar degeneration (FTLD). Although clinical symptomatology differs between the diseases, symptoms in the early stage may be unclear and can overlap \citep{McKhann2011, Gorelick2011, Rascovsky2011}. The current clinical criteria for AD and FTLD for example, which entail qualitative inspection of neuroimaging, fail to accurately differentiate the two diseases \citep{Harris2015}. Additionally, a young patient ($<$ 65 years old) with behavioural problems could have a differential diagnosis of dementia (i.e., behavioural phenotypes of FTLD or AD) or primary psychiatric disorder, as symptomatology overlaps substantially \citep{Kuruppu2013}. An accurate diagnosis of primary psychiatric disorder can be informative in such patients by suggesting  that progressive decline in the condition is not necessarily expected \citep{Ducharme2015}. For some specific diseases, measurements of proteins causing the underlying pathology have in the last decade shown high accuracy for diagnosis of the pathology. AD is a good example with blood-based biomarkers measuring phosphorylated-tau (P-tau), CSF biomarkers measuring Amyloid $\beta$, P-Tau and Tau, and PET imaging measuring Amyloid-$\beta$ and Tau. However, while highly promising, measurement of these proteins is not yet widely performed in clinical practice as blood-based biomarkers of AD are not widely available yet, {CSF biomarkers require an invasive lumbar puncture}, and PET imaging is too expensive and not sufficiently widely accessible to be done in each patient. Moreover, such markers of the underlying pathology are currently unavailable for other types of dementia. As an alternative, quantitative neuroimaging and other biomarkers, especially in combination with machine learning and large data sets, have shown to be beneficial in difficult cases of differential diagnosis \citep{Bron2017, Raamana2014}.

Another disorder where differential diagnosis is crucial is brain cancer. Diagnosis of brain tumours typically starts with the analysis of MRI brain data. A first diagnostic task is to differentiate between primary and secondary lesions. Primary lesions are tumours that originated from healthy brain cells, with glioma being the most common primary brain tumour type. Secondary lesions are metastases from tumours located elsewhere in the body, which may trigger very different care and treatment paths. Also the distinction between glioma and other less common malignant primary lesions such as lymphoma is relevant. Whereas neuroradiologists are trained to differentiate these different types of lesions, the large variation in appearance of tumours induces uncertainty in the differential diagnosis. Machine learning has been shown to be able to distinguish glioma from metastasis \cite{Chen2019} and lymphoma \cite{McAvoy2021} based on quantitative analysis of brain MRI, and may thus be used as a `second' reader supporting the radiologists. Once a diagnosis of cancer is established, a second task in differential diagnosis is the further subtyping of the lesion. While glioma is one of the deadliest forms of cancer \cite{Office2019}, there exist large differences in survival and treatment response between patients. These differences can be attributed to the glioma’s genetic and histological features, in particular the isocitrate dehydrogenase (IDH) mutation status, the 1p19q co-deletion status, MGMT promoter methylation status, and the tumour grade \cite{Dubbink2015, EckelPassow2015, Gessler2019}. These insights have led to classification guidelines by the World Health Organization (WHO) \cite{Louis2021}. In current clinical practice, these genetic and histological features are determined from tumour tissue after resection. However, there has been an increasing interest in complementary non-invasive alternatives that can provide the genetic and histological information before resection \cite{Zhou2018, Bi2019}. Also here, neuroradiologists can be trained to visually distinguish the subtypes based on MRI \cite{Smits2016, Delfanti2017}, but uncertainty often remains and the inherent subjectivity associated with visual inspection of subtle differences in appearance, by radiologists with varying levels of expertise, is undesirable. A large body of research has therefore focused on development of machine learning approaches to support MRI-based determination of genetic and histological features of glioma \cite{Sarkiss2019, Kocher2020, Gore2021, Singh2021}.

The third diagnostic task we address is \textbf{decision making for treatment}. This is relevant when multiple therapeutic options are available, such as for patients with stroke. Multiple treatment options for stroke exist such as thrombolytic medication and endovascular clot retrieval (mechanical thrombectomy). Since depending on the situation different treatments or their combination may be optimal, and since the costs per patient are rising, there is a real and urgent need for computer-aided diagnosis techniques to aid in the streamlined care of patients and individualised treatment decisions \citep{Kamal2018}. To enable early treatment of acute stroke, early and reliable diagnosis is required, which heavily relies on imaging. The vast majority of strokes are of ischemic origin, caused by a blood clot occluding an artery resulting in oxygen deprivation of the brain tissue supplied by this artery. Typical causes are large vessel occlusion with or without thrombus dislodgement (e.g. carotid stenosis) or a cardiac cause resulting in embolies (e.g. atrial fibrillation). The less common subtype is hemorrhagic stroke, which has substantially different aetiology and is often caused by hypertension. Without early treatment of stroke, prognosis is poor. Each minute without treatment leads to loss of an estimated 1.8 million neurons \citep{Knight-Greenfield2019}. Patients that enter the hospital with acute stroke symptoms often immediately undergo CT (or MR) scanning, even before detailed clinical evaluation of the patient \citep{Knight-Greenfield2019}. Imaging here has three roles in decision making for treatment: 1) rule out hemorrhagic stroke; 2) establish the exact cause and the extent of ischemic stroke and 3) determine a patient’s suitability for (intra-arterial) treatment \citep{El-Koussy2014, Mair2014}. Applications of machine learning for treatment decisions in stroke include identification of haemorrhage and early identification of imaging findings to determine the cause and extent of stroke and estimation of the time of onset. Time of onset is relevant since most current treatments aim for rapid reperfusion of ischemic tissue, either using intravenous thrombolytic medications or using endovascular techniques to mechanically remove the obstruction to blood flow, which should be performed within 4.5 hours of stroke onset \citep{Kamal2018}.

\subsection{Prediction} \label{sec:introPrediction}

Prediction or prognosis aims to understand the future `condition' of the patient, which can then be used for considering and planning therapeutic or lifestyle interventions proactively~\citep{Crous-Bou2017} that may slow the disease process or may reduce the risk for event recurrence. In addition, it can be used for effective patient management, for managing the expectations of patients and their caregivers~\citep{Mank2021}, as well as for patient selection in clinical trials \citep{Ezzati2020, Oxtoby2021}. We distinguish two main categories of prediction targets here: the natural disease course and patient outcomes after treatment.

\begin{floatbox}[H]
    \begin{nicebox}[Prediction]
        \label{box:prediction}
         Categories of prediction targets for which computer-aided models have or could have added value, with example brain disorders for which this is relevant as discussed in this chapter:
        \begin{itemize}
            \item \textbf{Natural disease course} Dementia, MS
	\item \textbf{Patient outcomes after treatment} MS, Brain cancer, Stroke
        \end{itemize}
    \end{nicebox}
\end{floatbox}

\textbf{Predicting the natural disease course}, i.e. the future progression of the disease and its symptoms in a subject, is clinically relevant as it can aid care planning and managing the expectations of patients and caregivers about their future quality of life, physical health and dependency \citep{Mank2020}. Additionally, in disorders where treatment options are limited, it would improve future clinical trials for new medication through identification of patients most likely to benefit from an effective treatment, i.e. those at early stages of disease who are likely to progress over the short-to-medium term (1-5 years) \citep{Marinescu2020}. 

In dementia, prediction is challenging because of disease heterogeneity, i.e. differences in symptoms between patients along the disease process. For example, a patient can either have typical AD with memory problems, or atypical AD with either language problems \citep{GornoTempini2008} or behavioural problems \citep{Ossenkoppele2021}. Moreover, patients with comparable brain atrophy may decline differently as the disease progresses, reflecting cognitive resilience due to genetic or lifestyle factors that may help to compensate for the level of atrophy \citep{Yao2020}. Lastly, a similar symptom in two patients could be resulting from different diseases altogether. For example, a patient with mild cognitive impairment (MCI) may have either early stage dementia or may have cognitive impairment due to a different cause such as older age, injury, or a virus such as SARS-CoV-2 \citep{Hampshire2021}. The latter, i.e. cognitive impairment due to non-degenerative disorders, is almost twice as prevalent as cognitive impairment due to dementia \citep{Plassman2011}. Here it is of interest to predict how the symptoms will develop over time for an individual; while patients without dementia may remain stable over time or even improve, the symptoms of patients with dementia typically worsen with time. Hence, the applications of machine learning in predicting the future course of dementia include: i) predicting if a patient with cognitive impairment patient will develop dementia \citep{vanMaurik2019}, ii) predicting when the patient will reach a clinical dementia stage (i.e. duration of the prodromal disease phase) \citep{Marinescu2020}, and iii) predicting the progression of biomarkers such as cognition and MRI measurements \citep{Koval2021, Kim2021}. 

In MS, especially in the early stages when patients experience clinical symptoms sporadically, prediction of the future disease course is highly relevant for care planning and expectation management. The early stage of MS, known as the relapsing-remitting phase, is characterised by sporadic inflammatory attacks on the neuronal protective coating called myelin. Over time, the recovery from these relapses becomes incomplete, resulting in permanent and progressive disability \citep{Weinshenker1994}. Because of this progressive nature and the variation between individuals, predicting the number of relapses and the time to permanent disability in a specific patient is highly important for care and treatment planning \citep{Brown2020}.

Next to prediction of the natural disease course, prediction of the future disease course after an intervention, i.e. \textbf{outcome prediction after treatment}, could be instrumental for planning of treatment and subsequent follow-up. This is of particular interest in MS where multiple treatment options are available. There are currently $21$ FDA-approved disease-modifying drugs available\citep{NationalMSS2021} that inhibit different aspects of pathological progression of MS mainly by immune modulation and sometimes through neuroprotection or remyelination. It is hence clinically highly relevant to choose the treatment option that an individual patient is expected to have most benefit from and to determine whether risks of second-line treatment are justified \citep{Stuhler2020}. The same holds for stroke in the post-acute phase, where prediction of patient outcomes after treatment based on imaging may play a role for choosing between available treatments such as medication and rehabilitation therapy \citep{Mair2014}. Here the focus is on the long term: reducing risk of recurrence and optimization of functioning. Computer-aided approaches can thus help in personalising the treatment for a patient. 

Predicting the outcomes after treatment is also of major interest for patients with brain tumours, and specifically in case of glioma where treatment response varies greatly across patients. Treatment usually consists of surgical resection followed by radiotherapy and/or chemotherapy. Almost invariably tumour recurrence or regrowth occurs, however, the question is when. In case of high-grade glioma (i.e., glioblastoma), tumour regrowth typically happens within a few months. In low-grade glioma, progression after treatment is often slower, and it may take years before any significant regrowth is detected; at some point however, malignant transformation (to a high-grade glioma) may occur, leading to accelerated regrowth. As discussed in Section \ref{sec:introDiagnosis}, computer-aided diagnosis methods can be used to identify the current tumour’s genetic and histological profile, which already provides important prognostic information. Beyond this example of computer-aided differential diagnosis, machine learning methods can contribute in different ways by directly predicting outcomes after treatment \cite{Kocher2020, Singh2021}. First, machine learning methods have shown promise to aid the differentiation between tumour progression and treatment related abnormalities (pseudoprogression, radiation necrosis) \cite{Jang2018, Kocher2020, Wang2020, Singh2021, Li2021}. Second, machine learning can be used to predict local relapse locations after radiotherapy, thus highlighting locations that should be targeted with a higher radiation dose, leading to personalised radiotherapy planning \cite{Rathore2018}. Third, a machine learning approach can predict local response to stereotactic radiosurgery of brain metastases, based on radiomics analysis of pretreatment MRI, where the outcome of interest (local tumour progression) was defined in terms of maximum axial diameter growth as measured on a follow-up scan \cite{Mouraviev2020}. Fourth, machine learning methods have been proposed for prediction of progression-free and overall survival, which aids care planning and managing the expectations of patients about their future \cite{Kickingereder2018, Sarkiss2019, Qiu2020, Singh2021}.  

\section{Method evaluation} \label{sec:method}
\subsection{State-of-the-art methodology for diagnosis and prediction} \label{sec:methodSOTA}

For \textbf{early diagnosis in dementia}, a large body of research has been published on classification of subjects into AD, mild cognitive impairment (MCI), and normal aging \citep{Wen2020, Rathore2017, Falahati2014}. Overall, classification methods show high performance for classification of AD patients and cognitively normal controls with an area under the receiver-operating characteristic curve (AUC) of $85-98\%$. Reported performances are somewhat lower for early diagnosis in patients with MCI, i.e. prediction of imminent conversion to AD (AUC: $62-82\%$). Dementia classification is usually based on clinical diagnosis as a reference standard for training and validation \citep{McKhann2011}, but biological diagnosis based on assessment of amyloid pathology with PET imaging or CSF  has been increasingly used over the last years \citep{Jack2018, Son2020}. Structural T1-weighted (T1w) MRI to quantify neuronal loss is the most commonly used biomarker, whereas the support vector machine (SVM) is the most commonly used classifier. For T1w, both voxel-based maps (e.g. voxel-based morphometry maps quantifying local gray matter density \citep{Kloppel2008}) and region-based features \citep{Magnin2009} have been frequently used. While using only region-based volumes may limit performance, combining those with regional shape and texture has been shown to perform competitively with using voxel-wise maps \citep{Cuingnet2011, Bron2014, Bron2015}. Using multimodal imaging such as FDG-PET or DTI in addition to structural MRI may have added value over structural MRI only, but limited data is available\citep{Zhang2011, Liu2018}. Following the trends and successes in medical image analysis and machine learning, neural network classifiers - convolutional neural networks (CNN) in particular - have increasingly been used since a few years \citep{Wen2020, Bron2021}, but have not been shown to significantly outperform conventional classifiers. In addition, data-driven disease progression models are being developed \cite{Oxtoby2017}, which do not rely on a-priori defined labels but instead derive disease progression in a data-driven way. 

Regarding \textbf{differential diagnosis in dementia} studies focus mostly on discriminating AD from other types of dementia. Differential diagnosis based on CSF and PET biomarkers of AD pathology has shown good performance for distinguishing AD from FTLD with sensitivities of 0.83 (p-tau/Amyloid-$\beta$ ratio from CSF) and 0.87 (amyloid PET) \citep{Rabinovici2011, Hellwig2019, Rivero-Santana2017}. In addition, machine learning approaches have been published based on either structural or multimodal MRI as region-wise or voxel-wise imaging features and generally SVM as a classifier, similar to those used for early diagnosis in dementia. These methods focused mostly on differential diagnosis of AD and FTLD and reported performances in the range of AUC=0.75-0.85\citep{Raamana2014, Moller2016, Bron2017, Bouts2018}. A few studies addressed differential diagnosis of AD and vascular dementia (VaD) \citep{Zheng2019} or multiclass differential diagnosis (5+ classes including AD, FTLD, VaD, dementia with lewy bodies and subjective cognitive decline) \citep{Tong2017, Morin2020}.

\begin{figure}[H]
\centering
\begin{subfigure}[b]{0.48\textwidth}
\includegraphics[width=1\textwidth]{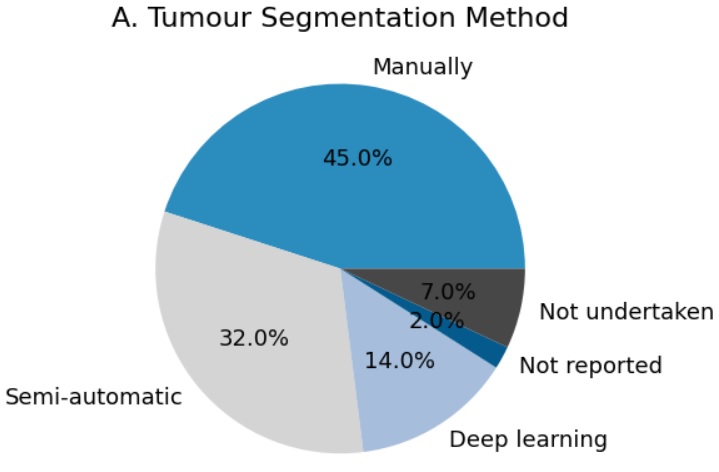}
\end{subfigure}
\begin{subfigure}[b]{0.48\textwidth}
\includegraphics[width=1\textwidth]{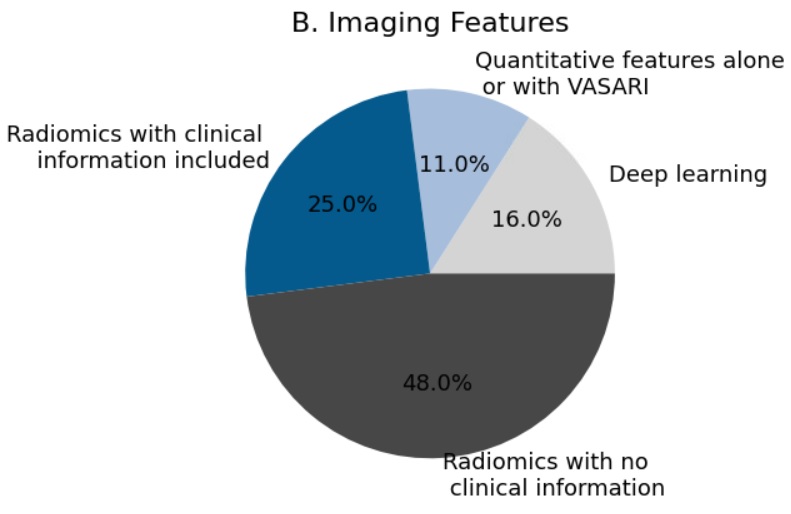}
\end{subfigure}
\begin{subfigure}[b]{0.48\textwidth}
\includegraphics[width=1\textwidth]{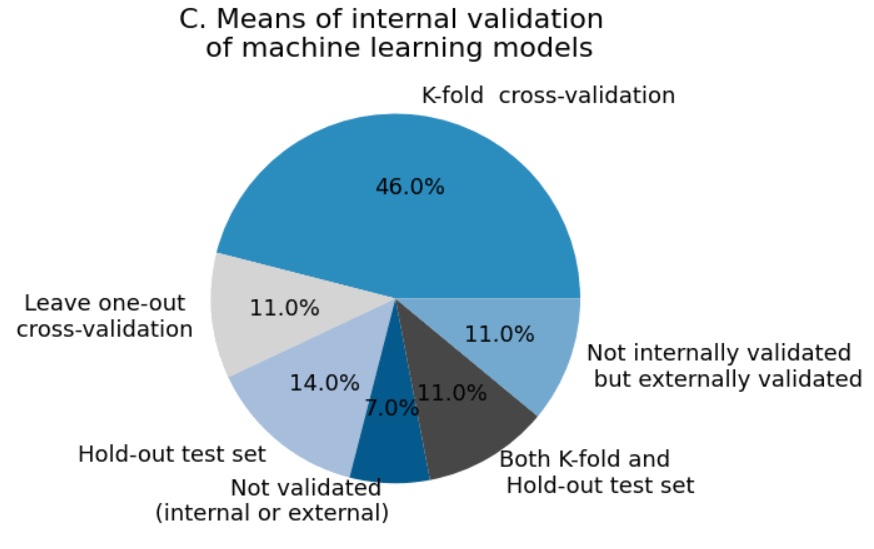}
\end{subfigure}
\begin{subfigure}[b]{0.48\textwidth}
\includegraphics[width=1\textwidth]{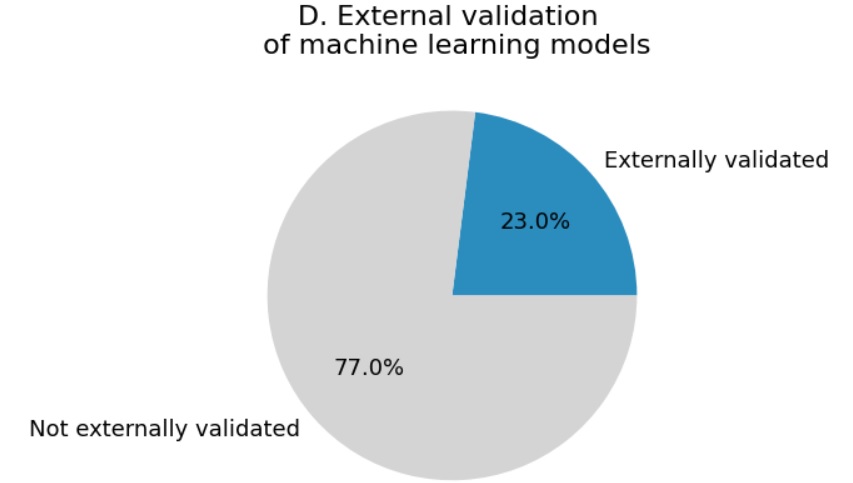}
\end{subfigure}
\caption{Summary of tumour segmentation methods \textbf{A}, types of imaging features \textbf{B}, means of internal validation \textbf{C}, and external validation \textbf{D} used by studies (n = 44) investigating machine learning models for predicting genetic subtypes of glioma. VASARI, Visually Accessible Rembrandt Imaging. Recreated from \cite{Jian2021}. Permission to reuse was kindly granted by the publishers.}\label{fig:jian2021}
\end{figure}

For \textbf{differential diagnosis in brain cancer}, numerous MRI-based machine learning approaches have been presented. These developments have partly been facilitated by the availability of several valuable public datasets, see for example the overviews in \cite{vandervoort2020, Menze2021}. Most literature is dedicated to glioma characterisation, which is therefore discussed in more detail here. Studies vary in the choice of input MRI sequences (T1w pre- and post-contrast, FLAIR, T2w, diffusion weighted imaging, perfusion weighted imaging, MR spectroscopy, APT CEST), the machine learning methodology (ranging from conventional radiomics approaches with hand-crafted features derived from manual tumour segmentations, to deep learning approaches that automatically segment the tumour), the classification target(s) (e.g. grade, IDH, 1p19q, and/or MGMT status), the selection of glioma subtypes on which the method is validated (e.g. only low-grade glioma, only high-grade glioma, or both), and the extent of validation performed (single train-test split, repeated cross-validation, internal versus external validation). A systematic review on the use of machine learning in neuro-oncology found four articles on glioma grading, and four articles on identifying genetic/molecular characteristics of glioma based on MRI \cite{Sarkiss2019}. Among those, only one study used convolutional neural networks as a machine learning tool - to predict 1p19q status in low-grade glioma \cite{Akkus2017}. A more recent systematic review identified 27 studies on glioma grading of which 6 used deep learning, and 48 studies on MRI-based estimation of genetic/molecular characteristics of which 8 used deep learning \cite{Buchlak2021}. Another recent review dedicated to machine learning approaches for MRI-based glioma characterisation found 12 studies on glioma grading of which 2 used deep learning, and 43 studies on molecular characterisation out of which 10 used deep learning \cite{Gore2021}. These numbers indicate a trend towards deep learning approaches as we see in the entire field, but with conventional machine learning approaches with pre-defined radiomics features still being used frequently. Regarding the performance, two recent systematic reviews performed a meta-analysis of studies on molecular characterisation of glioma. Jian et al. \cite{Jian2021} found a pooled sensitivity/specificity/AUC in the validation set of 0.85/0.83/0.90 for IDH status prediction (12 studies), and 0.70/0.72/0.75 for 1p19q status prediction (5 studies). For MGMT, sensitivities and specificities ranging from 0.70 to 0.88 were found in 3 studies reporting validation performance, not allowing a meta-analysis. Van Kempen et al. \cite{Kempen2021} reported a pooled AUC of 0.91 for IDH status prediction (7 studies), 0.75 for 1p19q status prediction (3 studies), and 0.87 for MGMT promoter status prediction (3 studies). Thus, while the studies applied somewhat different criteria for inclusion in the meta-analysis and used different statistical analysis methods, they obtained similar performance estimates . Whereas both meta-analyses suggest promising accuracy for MRI-based MGMT promoter status prediction based on the results reported in literature, a comprehensive evaluation of deep learning approaches for MGMT promoter status prediction on the BraTS2021 dataset \cite{Baid2021} yielded disappointing results, with AUCs ranging from 0.5 to 0.6 \cite{Saeed2022}. Also, the winning method of the BraTS2021 challenge achieved an AUC of 0.62 \cite{Bakas2021}, suggesting that MGMT promoter status prediction from MRI is a very difficult task. Both systematic reviews \cite{Jian2021, Kempen2021} also pointed out the low proportion of studies with external validation (10 out of 44 in \cite{Jian2021} and 12 out of 60 in \cite{Kempen2021}). Figure \ref{fig:jian2021}, {recreated based on} \cite{Jian2021}, shows a number of other insightful statistics on the methodologies found in literature. Finally, both reviews also identified machine learning methods aimed at predicting other, less frequently considered molecular targets, including ATRX, TERT, EGFR, P53, and PTEN, indicating the broad range of possible future research directions in this area.

Beyond glioma characterisation, other differential diagnosis problems in brain cancer are differentiation between glioma and lymphoma, between glioblastoma and metastasis, between different types of meningioma, and between glioma, meningioma and pituitary tumours \cite{Sarkiss2019, Kocher2020, Buchlak2021, Singh2021, Zegers2021}, with promising performances reported (AUC/accuracies around $90\%$). Of note, a recent study pointed out an important potential source of bias (the “Clever Hans effect”) in studies focused on differentiation between glioma, meningioma and pituitary tumours, due to implicit radiologist input in the selection of the 2D slices in a commonly used benchmark dataset \cite{Wallis2022}.

\textbf{For decision making in stroke}, different targets for machine learning based on imaging data have been identified, mostly focused at determining the cause and extent of stroke and to a lesser extent, on informing treatment decisions \citep{Kamal2018}. Regarding cause and extent of acute stroke, automatic lesion detection and identification of tissue-at-risk include the most important elements. These remain challenging as there is a lot of variation in lesion shape and location depending on time-from-symptom onset, vessel occlusion site, and collateral status \citep{Lee2017}. Machine learning methods for segmentation and detection are increasingly successful (see chapter 13). The step towards computer-aided diagnosis in stroke is also being taken using for example the CE-marked eASPECTS score \citep{Herweh2016}, which is a machine learning-based assessment of the Alberta Stroke Program Early Computed Tomography Score (ASPECTS). This system for scoring acute ischemic damage to the brain has shown to be a simple, reliable and strong predictor of functional outcome after stroke. Regarding treatment decisions, machine learning is used in several studies to determine whether a patient qualifies for a specific stroke treatment. For thrombolytic treatment, this qualification depends on time elapsed after symptom onset and treatment should be performed within 4.5 hours. For this application, methods are developed that provide a binary estimation of stroke onset time (i.e., more or less than 4.5 hours) based on either DWI and FLAIR \citep{Lee2020} or perfusion-weighted imaging (CT or MR) \citep{Ho2017}. Both approaches used a radiomics-like approach of feature extraction (e.g. intensity/gradient/texture based or using an autoencoder) followed by a machine learning classifier (support vector machine, random forest, and logistic regression). These machine learning methods had greater sensitivity than human readers using the standard procedure of DWI-FLAIR mismatch and comparable specificity. In addition, thrombolysis may cause the rare complication of symptomatic intracranial haemorrhage. Several machine learning methods have been developed to predict the risk of this complication achieving promising predictive performance, for example using a support vector machine classifiers based on CT data (AUC=0.74) \citep{Bentley2014}.

\begin{figure}[t]
\centering
\includegraphics[width=\textwidth]{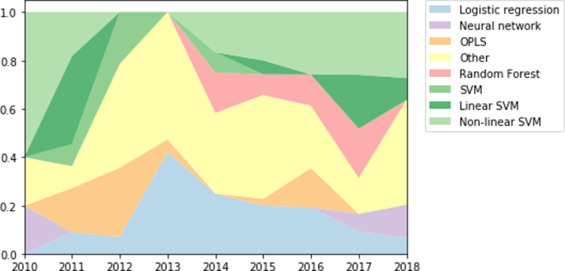}
\caption{Evolution with time of the use of various algorithms for predicting the progression of mild cognitive impairment. SVM with unknown kernel are simply noted as `SVM'. OPLS: orthogonal partial least square; SVM: support vector machine. Reproduced from \citep{Ansart2021}. Permission to reuse was kindly granted by the publishers.}
\label{fig:ansart2021}
\end{figure}

For \textbf{prediction of the future course of subjects at-risk of developing dementia}, there are three frequently used approaches for defining the prediction problem at hand. First, predicting whether the patient will develop dementia. In specific diseases, measurement of proteins causing underlying pathology has shown to be very promising to identify patients in a prodromal disease state. Here, prediction is performed either using univariate analysis or using logistic regression with few variables as input. Blood-based P-tau biomarker can predict incident AD within 4 years with an AUC of $0.78-0.83$ \citep{Palmqvist2021}, CSF biomarkers and PET images of amyloid $\beta$ and Tau can predict clinical progression of subjects in their prodromal AD state with an AUC of $0.94-0.96$ \citep{Hansson2018}. Alternatively, in the absence of pathology-specific markers, MRI and cognitive markers of a patient together with machine learning approaches have been used to predict AD with an AUC of $0.70-0.83$ \citep{Venkatraghavan2019b, Bron2021, Lin2018, Cui2011}. For a systematic review of the different machine learning methods developed for the purpose of predicting AD, see \citep{Ansart2021}. Support vector machines (SVM) and logistic regressions are the most used algorithms in the last decade (Fig. \ref{fig:ansart2021}). In FTD, where it is currently not possible to measure the pathological proteins in body fluids, prediction based on a combination of biomarkers that are non-specific to the underlying pathology is promising. This is demonstrated for example by ~\cite{vanderEnde2021}, who predicted disease onset in familial FTD based on unspecific blood-based and CSF-based biomarkers using a disease progression model and identified presymptomatic subjects that developed dementia in the near future with an AUC of $0.85$. 

Second, predicting the time for conversion to dementia. While the previous problem predicts a dichotomous output variable, here it involves predicting a continuous variable of time to dementia. Bilgel \textit{et al.} \citep{Bilgel2019} predicted time to AD dementia with a mean error of $< 1.5$ years. In the TADPOLE challenge, machine learning approaches to predict time for conversion to AD dementia of 33 participating teams have been assessed quantitatively~\cite{Marinescu2020}. Ansart \textit{et al.} \cite{Ansart2021} strongly favour predicting the exact time for conversion to dementia and argue against predicting converters within a given time interval (for example, within 3 years), because of the precision in the predictions. While this is indeed methodologically more elegant, the implications for clinical use and perception of patients regarding prediction precision and the inherent uncertainty remains to be established.

Third, prediction of disease markers could help to obtain insight into the clinical prognosis in an individual. Important disease markers are for example measures of global cognition (mini-mental state examination [MMSE] or Alzheimer’s disease assessment scale [ADAS] scores), or salient imaging markers (volume of the brain ventricles or longitudinal Tau protein accumulation). ADAS scores could not be reliably predicted by any participating team in the TADPOLE challenge~\citep{Marinescu2020}, but a recent disease progression model called \textit{AD course map} \citep{Koval2021} could predict ADAS scores (which is scored from 0 to 150) after 3 years with a mean absolute error of 7.6 points. \textit{AD course map} could also predict MMSE scores (which is scored from 0 to 30) after 3 years with a mean absolute error of 3.2 points. While these predictions used MRI as input, Tau PET was recently shown to be more predictive of future MMSE scores using linear mixed models \citep{Ossenkoppele2021b}. However, a thorough validation of this Tau PET-based prediction is lacking. Predicting salient imaging markers such as volume of the ventricles~\citep{Marinescu2020}, volume of the hippocampus \citep{Koval2021} or longitudinal Tau accumulation ~\cite{Leuzy2021} is a promising topic. Identifying the most clinically useful target to be predicted, the imaging modality that has the best cost-benefit ratio for prognosis of a patient, and the method that best predicts it are all important questions that still need answers in the future.

Most \textbf{prediction methods in MS} focus on predicting either physical disability, cognitive impairment, or treatment response {in imaging data} of an individual patient \citep{Kanber2019}. Physical disability as measured by expanded disability status scale (EDSS, range 0-10) has been the most commonly used predictor variable as recently used in \citep{Pellegrini2020, Roca2020}. An ensemble of classifiers consisting of convolutional neural networks, random forests, and manifold learning was reported to predict EDSS with a mean square error of 3.0 \citep{Roca2020}. Cognitive impairment has either been predicted as a global measure of cognition or as specific cognitive domains such as attention or working memory~\cite{Eijlers2018}. For predicting treatment response in MS, Signori \textit{et al.} \citep{Signori2015} used meta-analysis to identify subject characteristics that have higher treatment effects. In \citep{Eshaghi2021}, the authors used an unsupervised disease progression model to identify subtypes of progression pathways in MS and found in  post-hoc analysis that one of the subtypes predicted  better treatment effects. Current challenges in this evolving field of predicting treatment response in MS and future directions have been summarized in \citep{Gasperini2019}.

For the \textbf{prediction of patient outcomes after treatment of brain cancer}, most machine learning studies have focused on MRI-based prediction of progression-free survival or overall survival, which will therefore be discussed in more detail here. A systematic review by Sarkiss \textit{et al.} identified nine articles on survival prediction in glioma, and two on survival prediction for patients after stereotactic radiosurgery of brain metastases \cite{Sarkiss2019}. A more recent systematic review by Buchlak \textit{et al.} identified 17 studies on survival prediction with performance estimates (AUC or accuracy) mostly in the range 0.7-0.8 \cite{Buchlak2021}. Among those, only one study reported results of external validation, predicting overall survival of patients with low-grade glioma, and obtained an AUC of 0.71 with a model combining radiomics with non-imaging features including age, resection extent, grade, and IDH status \cite{Choi2020}. Random (survival) forests and support vector machines were most often used methods. One study used a CNN as a pre-trained feature extractor \cite{Lao2017}. Other recent approaches using CNNs to extract features that are subsequently combined with other factors into a final prognostic model include \cite{Nie2019, Han2020, Huang2021}. The 2017/2018 editions of the well-known BraTS challenges also included a task on overall survival prediction, with best teams obtaining accuracies around 0.6 in a three-class classification setting distinguishing short-, mid-, and long-survivors, \cite{Bakas2018}. Here, it was also pointed out that conventional machine learning methods outperformed deep learning methods, likely due to the limited size of available datasets for training.

Beyond MRI-based methods, methods using histopathology images and/or genomics data as input for the machine learning model are also considered in the literature on outcome prediction for glioma patients. In one of the pioneering studies on digital pathology images of glioma, better prognostication was obtained with deep learning when pathology images were combined with genetic markers (IDH, 1p19q) \cite{Mobadersany2018}. Preliminary work on so-called ‘radiopathomics’ in glioma is also available, supporting the notion that combining histology and radiology features improves prognostication (overall survival prediction) in glioma patients \cite{Rathore2019, Rathore2019b}.

\subsection{Benchmarks and challenges} \label{sec:methodBenchmark}
For 15 years, grand challenges have been organised in the biomedical image analysis research field. These are international benchmarks in competition form that have the goal of objectively comparing algorithms for a specific task on the same clinically representative data using the same evaluation protocol. In such challenges, the organisers supply reference data and evaluation measures on which researchers can evaluate their algorithms. Over the past years, the number and the impact of such grand challenges have increased \citep{Maier-Hein2018}. Also in the field of computer-aided diagnosis and prediction, such grand challenges have been organised. For example, in the dementia field, four challenges have been organised focusing on early diagnosis \citep{Bron2015, Allen2016, Sarica2018} and predicting the natural disease course \citep{Allen2016, Sarica2018, Marinescu2020}. In general, algorithms winning the challenges performed rigorous data pre-processing and combined a wide range of input features \citep{Bron2022}. In the field of brain cancer, the series of BraTS challenges has had a major impact \cite{Bakas2018, Baid2021}. These benchmarks are instrumental to gaining insight into successful approaches and their potential for use in clinical practice and clinical trials. 
 
\subsection{Open source software} \label{sec:methodOpen} 
Open source machine learning software such as Scikit Learn\footnote{\url{https://scikit-learn.org/}} and MONAI\footnote{\url{https://monai.io/}} have been fundamental to development of this field of research. More specifically for computer-aided diagnosis and prediction in brain diseases, dedicated platforms are available such as Clinica \citep{Routier2021}, NeuroPredict \citep{Raamana2018}, and PRoNTo \citep{Schrouff2013}. We also see a trend of researchers publishing their scripts and trained classifiers with their publications in order to promote reproducibility.

\section{Clinical Impact} \label{sec:impact}

There are multiple ways in which computer-aided diagnosis and prediction models can make an impact on clinical practice. Key areas of impact are in decision making for treatment and care, replacing invasive diagnostic procedures and patient selection for clinical trials. Here we will discuss to what extent these clinical needs are addressed by current methods.

First, the most direct impact is on decision making for treatment and care. This not only affects clinical care and treatment planning in patients with for example dementia, stroke, MS or brain cancer, but is also important for managing the expectations of patients and their caregivers. Although high performances are achieved for some related tasks such as dementia classification, validation of those results on external datasets and clinical cohorts is still very limited as well as knowledge on the robustness of the methods. For other applications, there is still room for performance improvement, and key factors in achieving that would be the combination of multi-modal input and the availability of more well-maintained and large-scale datasets for training and evaluation. In general, there is room for improvement in how well real clinical questions are addressed by current methodology. Second, machine learning models can have an impact by replacing invasive diagnostic procedures. This is especially relevant in brain cancer, where machine learning techniques based on imaging data are developed to predict for example genetic mutation status or tumour grade, thereby avoiding or reducing the need for biopsies \citep{Zhou2018, Bi2019}. As a motivating example, MRI-based prediction of MGMT methylation status could be beneficial to guide treatment decisions. This is supported by findings from a population-based study assessing survival in 131 patients with radiological diagnosis of glioblastoma who did not undergo surgery and thus lacked (histological or molecular) tissue-based verification of the diagnosis \cite{Werlenius2020}. While patients without treatment had extremely poor prognosis with median survival of 3.6 months, those who received upfront temozolomide treatment did significantly better (with median survival of 6.8 months). Since the response to temozolomide is known to be highly dependent on the MGMT status, MRI-based prediction of MGMT status could give insight into which patients would benefit from treatment avoiding the need for biopsies in patients to frail for tumour biopsy. Third, patient selection for clinical trials is relevant in diseases where no to limited options for treatment exist, such as dementia, or diseases where existing treatments are suboptimal for some patients, such as MS. This can boost the power of trials by enrolling for example individuals who are more likely to progress based on prediction models. Several pilot studies demonstrated the added value of machine learning models to select a subgroup of participants to increase sensitivity to the treatment using phase III trial data (e.g. for  Alzheimer’s disease treatment using donepezil or semagacestat) \citep{Ezzati2020, Oxtoby2021}. This will ultimately reduce size, duration and cost of clinical trials.

The number of published methods is not evenly distributed over tasks. While many methods have been published on for example the classification of Alzheimer’s disease patients versus controls, much fewer publications exist on differential diagnosis in dementia. In addition, there seems to be a mismatch in some applications between published classification methods and clinical needs, e.g. the clinically relevant problem of early diagnosis does not directly translate to the frequently studied classification task of established Alzheimer’s disease versus healthy controls, but would instead require separation of early disease stage Alzheimer’s disease patients from those that have cognitive complaints but not dementia.  

Several approved machine learning products to assist diagnosis and prediction are making their way into clinical practice, in particular in the imaging domain. Van Leeuwen et al. evaluated 100 commercially available products for AI in radiology, of which 38 are related to brain diseases \citep{VanLeeuwen2021}. These include mostly segmentation, quantification and normative comparison for neurodegenerative diseases and detection of lesions for stroke and oncology. Most methods generate a sample radiologist report which can be inspected and modified. In dementia, for example, 17 reporting tools that use automated brain MRI segmentation software and normative reference data for single-subject comparison are regulatory approved for use in the memory clinic \citep{Pemberton2021}. 

One of these is Quantib ND (Quantib BV, Rotterdam, the Netherlands)\footnote{\url{quantib.com/solutions/quantib-nd}}, which is an approved commercial software that performs automatic segmentation into 20 brain regions as well as normative volumetry reference curves based on data of ~5000 subjects from a population based cohort. While Quantib ND and most other available tools use machine learning for brain segmentation, their output is not a diagnostic label produced by a machine learning algorithm. Another approved software, cDSI (Combinostics, Tampere, Finland)\footnote{\url{combinostics.com/cdsi}}, does output diagnostic labels as confidence scores in addition to segmentation and normative volumetry based on MRI. It uses univariate machine learning to normalise individual biomarkers of different modalities based on reference values of patient and control groups, colour-codes these biomarkers to improve visualisation of large-data datasets and combines confidence scores based on individual biomarkers into one score \citep{Mattila2012a, Mattila2012b}. While cDSI is a machine learning tool for computer-aided diagnosis and prognosis, it does not exploit the power of machine learning to detect complex patterns in high dimensional data but rather focuses on visualisation and interpretability. Diagnosis and prediction algorithms that map high dimensional input, i.e. images and other clinical data, to an outcome measure using machine learning have not yet made their way into clinical practice.  

\section{Roadmap for clinical translation} \label{sec:impactRoadmap}

There are numerous challenges for clinical translation of computer-aided diagnosis and prediction methods. Some key items that should be on the roadmap for translation relate to large and standardised datasets, to technical and clinical validation, to interpretability by clinicians and patients, and to practical issues related to implementation. In this section, we will discuss these requirements and related developments and initiatives.

The first requirement for translation are \textbf{large and standardised datasets}. For a few brain disorders, one or multiple large datasets (i.e. up to 2500 participants) are available to train machine learning algorithms for diagnosis and prediction tasks, facilitated by large multicenter initiatives such as the Alzheimer’s disease Neuroimaging initiative (ADNI) or the Parkinson’s progression markers initiative (PPMI). For validation in other cohorts and for development of algorithms in other diseases, there is only limited data available and a need for more (well-annotated) data exists. {In particular there is a need for validation data that reflect the reality of clinical routine with no to limited data harmonization and large variation in imaging protocols and data quality. }Setting up such large-scale datasets is complex due to various reasons including obstacles in inter-institutional data sharing and a lack of funding for collection, curation and labelling of data. To overcome these challenges, developments in research software and infrastructure may provide a solution by sharing easily reproducible algorithms rather than the data. Wrapping an algorithm in a container (e.g. Docker\footnote{\url{www.docker.com}}, Singularity \citep{Kurtzer2017}) and applying the algorithms locally to the data (at one site or multiple sites in a federated approach) enables method validation on large sets of data within the confines of the local institute’s firewalls. Such an approach could be also used for enabling training on larger datasets (i.e. federated learning~\citep{Santiago2020}). Standardisation of the data is important for eventual translation as it enables researchers to combine multiple datasets for development and validation of machine learning methods for diagnosis and prediction. Such standardisation entails both data collection (e.g. diagnostic criteria, protocols for image acquisition and clinical tests) and data organisation (e.g. through open source standards and platforms for data storage such as the Brain Imaging Data Structure (BIDS) and the Extensible Imaging Archive Toolkit (XNAT)).

Second, \textbf{technical and clinical validation} is a key focus area on the roadmap for translation. In the field of radiology, the quantitative neuroradiology initiative (QNI) framework has been developed as a model framework for translation defining the technical and clinical validation necessary to embed automated software into the clinical workflow \citep{Goodkin2019}. Based on this framework, \cite{Pemberton2021} reviewed the published evidence regarding commercial automated volumetric MRI tools for dementia diagnosis. For the 17 products identified, 11 companies have published some form of technical validation on their methods, but only 4 have published clinical validation in a dementia population. They concluded that there is a significant evidence gap in the literature regarding clinical validation and in-use evaluation. Whereas this review only addressed image volumetry in dementia, these findings likely extend to other brain diseases, applications and modalities. Hence, there is a need for both retrospective and prospective studies validating algorithms in a clinical setting. In addition, performance metrics used in validation studies should aim to capture real clinical applicability and address different aspects of the reliability of an algorithm, including accuracy, uncertainty estimation, reproducibility and generalizability to other data. Standards for validation and reporting are provided by guidelines such as STARD-AI \citep{Sounderajah2021} and TRIPOD-AI\footnote{\url{osf.io/zyacb}}. 

A third key item for clinical translation is \textbf{interpretability} by end-users such as clinicians and patients. As clinicians have responsibility for the decisions related to care and treatment, they should have trust in a computer-aided diagnosis or prediction system and understand its outputs to an extent that they can rely on them for decision making and explanation to a patient. Performance metrics should aim to capture real clinical applicability and be understandable to intended users \citep{Kelly2019}. High validation performance is important for building trust in methods, but not sufficient by itself, since performance may reduce in individual cases because of unaccounted inter-individual such as comorbidities or population differences such as MRI scan protocol. Therefore, apart from model accuracy, relevant questions for interpretation are for example: Is the model suitable for the data of this patient? What features contribute to the machine learning decision for this patient? How certain is the decision for this patient and can the algorithm know when it is uncertain about an individual’s decision? Such questions are important and methods should be designed and implemented in a way that facilitates answers to such questions. This could be obtained by using interpretability methods on top of `black box’ machine learning models or directly by using interpretable models. For the first category, many methods have been developed based on model weight visualisation, feature map visualisation, back-propagation methods or perturbation of inputs (see also Chapter 22). For interpretable models, an example in the field of computer-aided diagnosis and prognosis are disease progression models\citep{Young2014, Venkatraghavan2019}. These data-driven models are designed specifically for neurodegenerative diseases and explain their decisions based on their estimate of the natural progression of the disease in the cohort (see also Chapter 17). 

As a final key item, we will discuss \textbf{implementation feasibility}. For machine learning models to be actually used in practice, it is essential that models and reporting are integrated into the clinical workflow and that the sending and processing of clinical data and receiving results is fully automated. Current commercial products for automatic volumetry in dementia all reported to have implemented an integration with radiology systems and the clinical workflow. While validation of the workflows is limited \citep{Pemberton2021}, this does support the feasibility for machine learning in clinical practice. While these products integrate with the radiological workflow, a key challenge for the clinical translation of algorithms that use non-imaging clinical data (such as cognitive scores) as input is to also integrate with the clinical workflow of multi-disciplinary diagnosis.

\section{Final summary and conclusion} \label{sec:conclusion}

Computer-aided diagnosis and prediction of brain disorders is an important research area, with a wide variety of applications. While typically for these applications generic machine learning methods are used, domain knowledge of these brain disorders is crucial for selecting novel clinically relevant applications as well as for making domain specific methodological improvements. Regarding diagnosis, clinical challenges are in early diagnosis of dementia and MS, differential diagnosis of dementia and brain cancer, and decision making for treatment in stroke. Regarding prediction, challenges are in the prediction of the natural disease course in dementia and MS, and the prediction of patient outcomes after treatment in stroke, brain cancer, and MS. {Even though the disorders on which we focused are important avenues for impact, computer aided diagnosis and prognosis would also be extremely useful in other disorders such as movement disorders for predicting response to treatment and side-effects, epilepsy for predicting response to epilepsy surgery, psychiatric disorders where diagnosis can be particularly difficult.}

Key areas of impact are in 1) decision making for treatment and care in patients with dementia, stroke, MS or brain cancer, 2) replacing invasive diagnostic procedures in brain cancer, and 3) patient selection for clinical trials in dementia and MS. While the first AI methods are making their way to clinical practice, diagnosis and prediction algorithms that map high dimensional input, i.e. images and other clinical data, to an outcome measure using machine learning are not yet clinically available. To enable translation, major items on the roadmap relate to the availability of large and standardised datasets and technical and clinical validation of the developed machine learning methods. In addition, other important aspects are interpretability of the results by clinicians and patients, optimisation of the diagnostic or treatment workflow in the clinic, and  other practical issues related to implementation. 

With this chapter, we aimed to provide a comprehensive overview, bringing together the clinical context of representative use cases of diagnosis and prediction in brain disorders and their state-of-the-art computer-aided methods. Future research should focus on bridging the identified gaps between clinical needs and the solutions brought by machine learning, to further improve decision making, treatment, and care in brain diseases.

\section*{Acknowledgements} \label{sec:ackn}

V. Venkatraghavan is supported by JPND-funded E-DADS project \\(ZonMW project $\#733051106$). W.J. Niessen and E.E. Bron are supported by Medical Delta Diagnostics 3.0: Dementia and Stroke. E.E. Bron acknowledges support from the Netherlands CardioVascular Research Initiative (Heart-Brain Connection: CVON2012-06, CVON2018-28). 

\section*{Conflict of interest} \label{sec:conflict}

Quantib BV is a spin-off company of Erasmus MC. W.J.N. is cofounder, part-time Chief Scientific Officer, and stock holder of Quantib BV. S.R.V, D.B., M.S., S.K. and E.E.B. are affiliated to Erasmus MC but have no personal relationships with or financial interest in Quantib BV. F.B. is a consultant to Combinostics.

\bibliographystyle{spbasic}
\bibliography{references}

\end{document}

%% file: template.tex
\usepackage[noblocks]{authblk} 
\usepackage[numbers]{natbib}
\usepackage{graphicx}
\usepackage{fancyhdr}
\usepackage[colorlinks=true, linkcolor=MidnightBlue, urlcolor=MidnightBlue, citecolor=PineGreen]{hyperref}
\usepackage[tableposition=top, justification=justified, textfont=it]{caption}
\usepackage{titlesec}
\usepackage{multicol}
\usepackage{afterpage}
\usepackage{lipsum}
\usepackage[dvipsnames, x11names]{xcolor}
\usepackage[inner=4cm, outer=4cm, top=2.8cm, bottom=2.8cm, marginparsep=0cm, marginparwidth=0cm, includehead, includefoot]{geometry}
\usepackage{setspace}
\usepackage{array}
\usepackage{bm}
\usepackage{tikz}
\usepackage{float}
\usepackage[framemethod]{mdframed}
\usepackage{amsmath}
\usepackage{amssymb}

\titleformat{\section}{\Large \bfseries \centering \scshape}{\thesection.}{0.3em}{}[{\titlerule[0.5pt]}]

\definecolor{shadecolor}{RGB}{230,230,230}
\newcommand{\mybox}[1]{\par\noindent\colorbox{shadecolor}
{\parbox{\dimexpr\textwidth-2\fboxsep\relax}{#1}}}

\titleformat{\subsection}{\large \bfseries \mybox}{\thesubsection}{1em}{}

\titleformat{\subsubsection}{\itshape}{\thesubsubsection.}{0.3em}{}

\renewenvironment{abstract}
{\vskip 2.5ex {\large\bf\noindent Abstract}\vspace{0.7ex} \\ %
  \bgroup\noindent\ignorespaces}%
{\par\egroup\vskip 2.5ex}

\newenvironment{keywords}
{\bgroup\leftskip 20pt\rightskip 20pt \small\noindent{\bf Keywords:} }%
{\par\egroup\vskip 10ex}

\fancypagestyle{firstpage}{%
  
  \fancyhead[]{}
  \fancyfoot[C]{}
  \fancyfoot[L]{\footnotesize To appear in \\
  O. Colliot (Ed.), \textit{Machine Learning for Brain Disorders}, Springer\\
  }

}

\fancypagestyle{otherpages}{%
  
  \fancyhead[LE]{\thepage}
  \fancyhead[RE]{\runningauthor}
  \fancyhead[LO]{\runningheadtitle}
  \fancyhead[RO]{\thepage}
  \fancyfoot[L]{\footnotesize  \textit{Machine Learning for Brain Disorders}, Chapter~\chapternumber}
  \fancyfoot[C]{}
}

\makeatletter
\renewcommand{\maketitle}{\bgroup\setlength{\parindent}{0pt}

\begin{flushright}
  \color{MidnightBlue}
  \textbf{\LARGE Chapter~\chapternumber}
\end{flushright}

\vspace{0.3in}

\begin{flushleft}
    \setstretch{2.0} 
    \textbf{\color{MidnightBlue}\huge\@title}
\end{flushleft}

\vspace{0.15in}

\begin{flushleft}
    \textbf{\bfseries \large\@author}
\end{flushleft}\egroup
}

\makeatother

\DeclareCaptionLabelFormat{labelformat}{\color{MidnightBlue}\bfseries #1 #2}
\renewcommand{\bibpreamble}{\scriptsize \begin{multicols}{2}}
\renewcommand{\bibpostamble}{\end{multicols}}

\mdfsetup{skipabove=\topskip, skipbelow=\topskip}

\newcounter{nicebox}
\newenvironment{nicebox}[1][]{%
    \refstepcounter{nicebox}%
    \ifstrempty{#1}%
    {\mdfsetup{%
        frametitle={%
            \tikz[baseline=(current bounding box.east),outer sep=0pt]
            \node[anchor=east,rectangle,fill=blue!20]
            {\strut Theorem~\thetheo};}}
    }%
    {\mdfsetup{%
        frametitle={%
            \tikz[baseline=(current bounding box.east),outer sep=0pt]
            \node[anchor=east,rectangle,fill=blue!20]
            {\strut Box~\thenicebox:~#1};}}%
    }%
    \mdfsetup{innertopmargin=10pt,linecolor=blue!20, linewidth=2pt,topline=true, frametitleaboveskip=\dimexpr-\ht\strutbox\relax,}
    \begin{mdframed}[]\relax%
    }{\end{mdframed}}

\newfloat{floatbox}{thp}{lop}
\floatname{floatbox}{Box}

\DeclareMathAlphabet{\mathsfit}{\encodingdefault}{\sfdefault}{m}{sl}
\SetMathAlphabet{\mathsfit}{bold}{\encodingdefault}{\sfdefault}{bx}{n}